\documentclass[10pt,twocolumn,letterpaper]{article}

\usepackage[pagenumbers]{cvpr} % To force page numbers, e.g. for an arXiv version

\usepackage{graphicx}
\usepackage{subcaption}
\usepackage{amsmath}
\usepackage{amssymb}
\usepackage{booktabs}

\usepackage[linesnumbered,ruled]{algorithm2e}
\usepackage{multirow}
\SetKwInput{KwInput}{Input}                % Set the Input
\SetKwInput{KwOutput}{Output}              % set the Output
\SetKwComment{Comment}{/* }{ */}

\usepackage[pagebackref,breaklinks,colorlinks]{hyperref}

\usepackage[capitalize]{cleveref}
\crefname{section}{Sec.}{Secs.}
\Crefname{section}{Section}{Sections}
\Crefname{table}{Table}{Tables}
\crefname{table}{Tab.}{Tabs.}

\begin{document}

\title{On Steering Multi-Annotations per Sample for Multi-Task Learning}

\author{Yuanze Li\\
Harbin Institute of Technology\\
{\tt\small sqlyz@hit.edu.cn}
\and
Yiwen Guo\\
\\
{\tt\small guoyiwen89@gmail.com}
\and
Qizhang Li\\
Harbin Institute of Technology\\
{\tt\small liqizhang95@gmail.com}
\and
Hongzhi Zhang\\
Harbin Institute of Technology\\
{\tt\small zhanghz0451@gmail.com}
\and
Wangmeng Zuo\\
Harbin Institute of Technology\\
{\tt\small wmzuo@hit.edu.cn}
}
\maketitle

%%%%%%%%% ABSTRACT
\begin{abstract}
The study of multi-task learning has drawn great attention from the community. Despite the remarkable progress, the challenge of optimally learning different tasks simultaneously remains to be explored. Previous works attempt to modify the gradients from different tasks. Yet these methods give a subjective assumption of the relationship between tasks, and the modified gradient may be less accurate. In this paper, we introduce Stochastic Task Allocation~(STA), a mechanism that addresses this issue by a task allocation approach, in which each sample is randomly allocated a subset of tasks. For further progress, we propose Interleaved Stochastic Task Allocation~(ISTA) to iteratively allocate all tasks to each example during several consecutive iterations.
We evaluate STA and ISTA on various datasets and applications: NYUv2, Cityscapes, and COCO for scene understanding and instance segmentation. Our experiments show both STA and ISTA outperform current state-of-the-art methods. The code will be available.
\end{abstract}

%%%%%%%%% BODY TEXT
\section{Introduction}
\label{sec:intro}
Thanks to the development of deep learning, more complex problems which consist of multiple tasks should be considered\cite{wu2021yolop, li2021value, zamir2018taskonomy}. For example, autonomous driving requires processing many tasks simultaneously, \eg, object detection, instance segmentation, and distance estimation, in order to self-locate itself and safely navigate in an unstructured environment\cite{chowdhuri2019multinet, phillips2021joint_loc_perp}. Simply training a model for each task is hard to satisfy the requirement for some reasons. On the one hand, the common sense between tasks can be utilized to reduce test errors~\cite{vandenhende2020mti, papnet, padnet} and improve robustness~\cite{safarani2021monkey, mao2020mtl_adv}. On the other hand, it requires too expensive computation cost when inference~\cite{misra2016cross, liu2019mtan}. 

Multi-task learning~(MTL) provides a solution that aims to train a network on all tasks simultaneously and discover a shared representation that is robust to various tasks. It enables a unified system to efficiently produce multiple predictions for different tasks at once and achieves better performance than solving tasks individually~\cite{chowdhuri2019multinet, phillips2021joint_loc_perp,fakhry2021mtlcost, vandenhende2021survey}.
Deep multi-task learning has achieved great success in several applications, including reinforcement learning~\cite{sharma2017rl_taskschedule, gradsurgery, hessel2019rl_popart}, computer vision~\cite{kendall2018multi,liu2019mtan,vandenhende2020mti, he2017mask, zamir2018taskonomy}, and natural language processing~\cite{liu2019nlp1, liu2019nlp2}. 

However, jointly learning multiple tasks results in a difficult optimization problem. Some works~\cite{guo2018dtp, kendall2018multi, chen2018gradnorm} consider that there is \textit{imbalance} between tasks where a task dominates the training process resulting in the stagnant of other tasks. In the context, related works always use a weighted sum of task losses as the multi-task loss and find ways to choose optimal weights for the imbalance problem, like ~\cite{du2018aux_weight, liu2019loss_weights1, crawshaw2021slaw}.
At the same time, multiple losses of tasks produce different gradients which may have opposite directions and eliminate each other when combining. This effect, known as \textit{gradient conflicts}, is considered as one of the main issues in multi-task optimization~\cite{gradsurgery,javaloy2021rotograd}. 
Prior works~\cite{lopez2017GEM, chaudhry2018AGEM} provide gradient replacement as a solution that replace gradients that conflicts heavily with a new version that has no conflicts, so that they may not be canceled when added together. 
This replacement only takes conflicts into accounts but the final gradients may be not enough accurate for the optimization of individual tasks. 

\begin{figure}[ht]
    \centering
    \includegraphics[width=\linewidth]{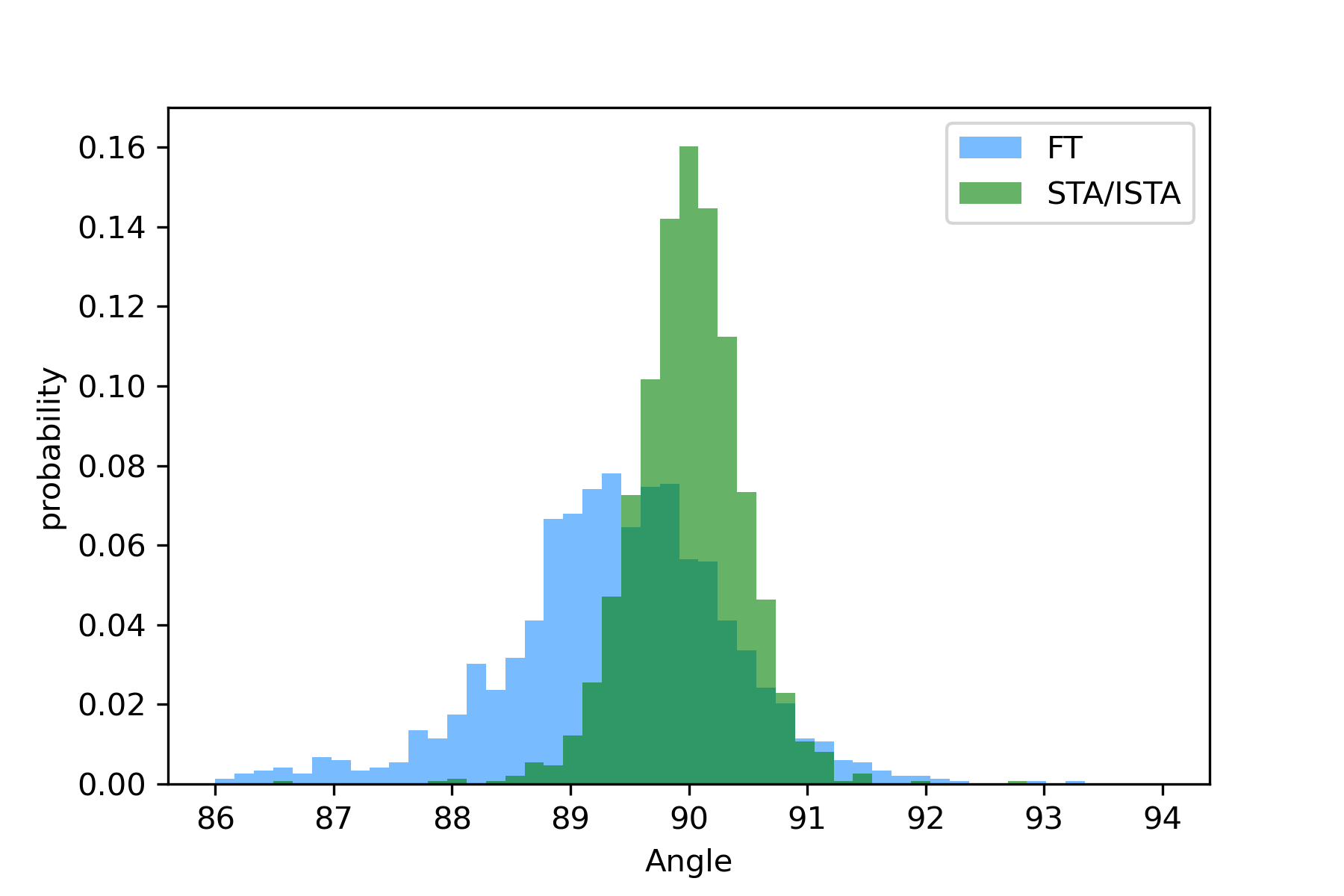}
    \caption{The distribution of angles between semantic segmentation and depth regression on Cityscapes. We use the same model to produce gradients and summary their angles while the model do not update. To stimulate training, we use same setting in \cref{table:city1} experiments, including batch size and initial model. We do static over 2 epochs. It is obvious that STA explicitly regularizes the angle between tasks around 90 degree. }
    \label{fig:cos_sa}
\end{figure}

This work focuses on \textit{gradient conflicts} and tries to avoid modifying gradients explicitly. We suppose that training an example with full tasks may be one of the main reason for gradient conflicts. With this hypothesis, we propose a novel method to soften gradient conflicts by randomly allocating partial tasks for each training example. We call it Stochastic Task Allocation~(STA). 
In \cref{fig:cos_sa}, we empirically show that STA changes the direction of per-task gradients implicitly where the distribution of gradient angles becomes more concentrated and orthogonal. Intuitively, more orthogonal gradients mean fewer conflicts and competitions in sharing nets. In related works~\cite{suteu2019cosreg, zhao2018unrelated_module}, they make the same claim that this trend is beneficial for the multi-task model.
We evaluate STA on NYUv2\cite{silberman2012nyuv2} and Cityscapes\cite{cordts2016cityscapes}. The results empirically show its superior performance than prior works. Additionally, we also apply STA on Instance Segmentation with Mask R-CNN~\cite{he2017mask} and achieve better performance on Cityscapes and COCO\cite{lin2014coco}.

\begin{figure}[ht]
    \centering
    \includegraphics[width=\linewidth]{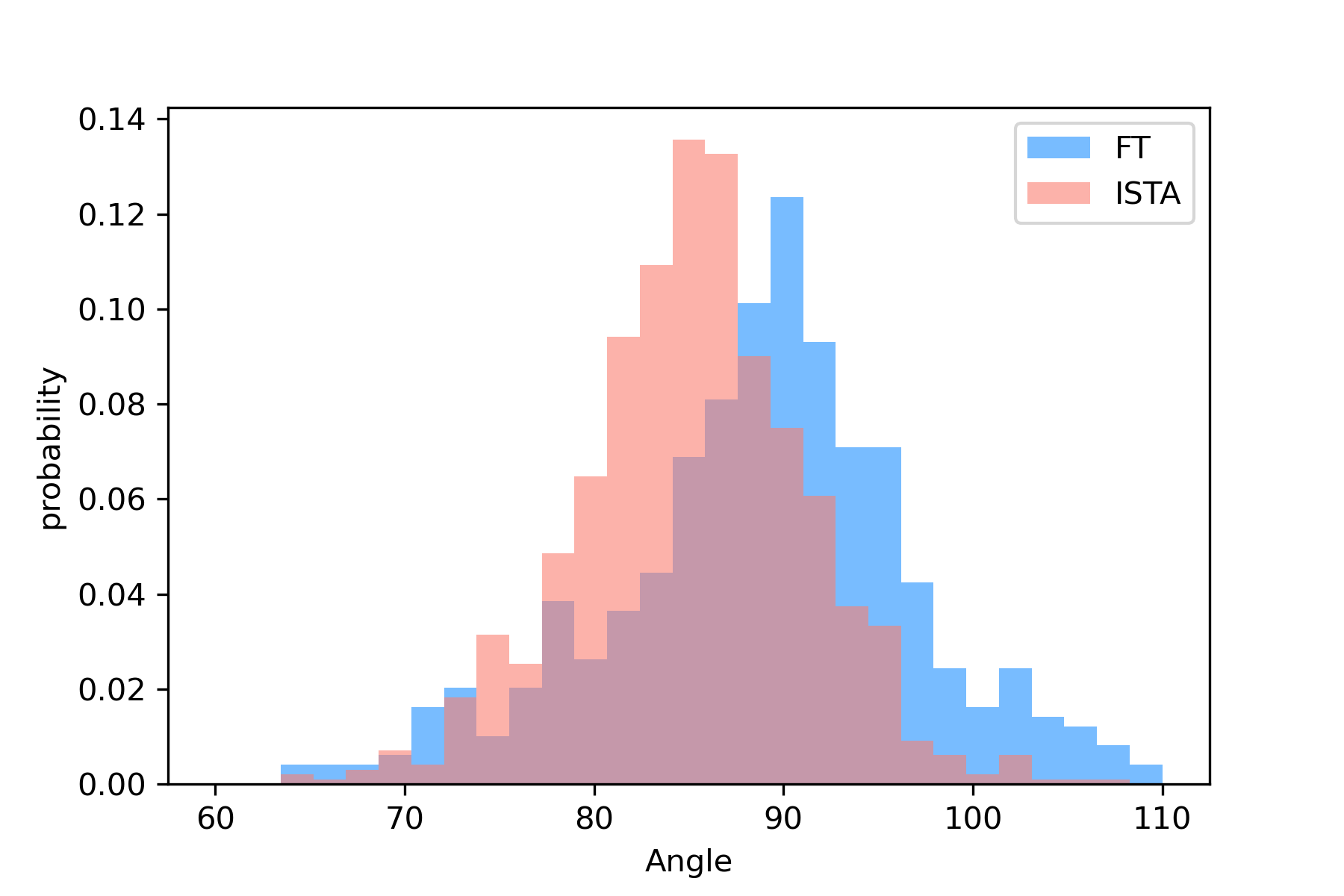}
    \caption{The distribution of gradient angles between steps on Cityscapes. We use a same model and summary the gradient angle between every two steps over 2 epochs, in order to illustrate the higher similarity between ISTA steps comparing to FT. Note that the model is not updating during the process.}
    \label{fig:cos_saita}
\end{figure}

\begin{figure*}
    \centering
    \includegraphics[width=0.9\linewidth]{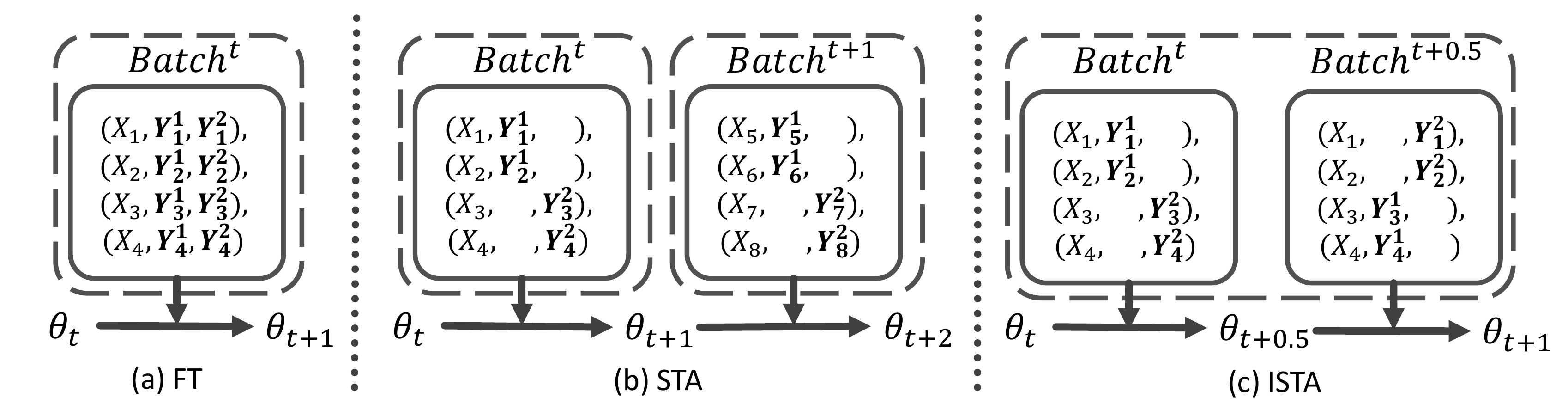}
    \caption{(a)~Baseline methods, train one example by Full Tasks at once. (b)~Our STA method, for each example, it stochastically allocates partial tasks~(only one task here) for each example. (c)~Our ISTA method contains several sub-steps. In sub-step $t$, it randomly allocate part of task for each example: Task 1 for $X_1$, Task 2 for $X_3$. In sub-step $t+0.5$, it allocate the rest task, Task 2 for $X_1$ and Task 1 for $X_3$ respectively. Here we take number of tasks $T=2$ as an example in order to clearly illustrate our STA and ISTA.}
    \label{fig:overview}
\end{figure*}

Further, we notice that it is inefficient to only utilize partial annotations in multi-task datasets. We propose Interleaved Stochastic Task Allocation~(ISTA) which make full use of all task information for each example, by progressively allocating all tasks over several steps until all task have been used. We give a brief overview to visualize the sampling process in \cref{fig:overview}.
We show that the gradients between ISTA steps are more correlated: they have a higher probability to attain smaller angles, as shown in \cref{fig:cos_saita}. It indicates that the interleaved allocation help to redress the multi-task information thanks to the use of rest tasks, while do not interrupt the training with more similar gradients. Moreover, we show empirically that ISTA can outperform STA on different architectures and datasets in \cref{sec:ablation}.

To sum up, the main contribution of this work includes:
\begin {itemize}
   \vspace*{-2.5mm}
   \item  A task sampling method, STA, is proposed to relief gradient conflicts problem by randomly allocating partial task for each example.
   \vspace*{-2.5mm}
   \item  A variance of STA, ISTA is committed to improving the data efficiency of STA by consecutively task allocation.
   \vspace*{-2.5mm}
   \item Extensive experiments on NYUv2 and Cityscapes show that our methods perform superior to the state-of-the-art methods of both task weighting and gradient replacement.
\end {itemize}

%------------------------------------------------------------------------
\section{Related Works}

In multi-task model, the optimization of different tasks is imbalance: there might be part of tasks dominating the training process while other tasks are inadequate to learn well. 
To solve this problem, some works focus on finding optimal weights between various classification and regression losses which have totally different scales and optimization properties. Uncertainty Weight(UW)~\cite{kendall2018multi} assumes that the optimal weights during training are related to the magnitude of tasks' noises, so it learns the homoscedastic uncertainty to form the optimal weights.
Different from uncertainty weights, GradNorm~\cite{chen2018gradnorm} uses gradient norms as the indicator and proposes an auxiliary loss to penalize tasks with higher gradient norms by lower weights.
Dynamic Task Priority~\cite{guo2018dtp} is another classical work to solve imbalance. It uses the metrics of current batch as the indicator and an exponential form to weight losses, like focal loss~\cite{lin2017focal}.

Prior works observe that multi-task training suffers from low similarities between per-task gradients so that gradients eliminate each other when being applied to the network.
A common way to soften conflicts is gradient replacement which modifies gradients directly. PCGrad\cite{gradsurgery} defines the gradient conflict as the obtuse angle between per-task gradients. While the conflicts happen, PCGrad tries to project one gradient onto the normal plane of another gradient. As a result, PCGrad successfully reduces the included angle below 90 degree. However, there is no guarantee that the projected gradients point to some optima of corresponding tasks. Our method, STA, does not explicitly change gradient direction but utilizes sampling to reduce gradient conflicts. 
RotoGrad\cite{javaloy2021rotograd} adds task-specific projection modules whose parameters are learnable. These modules produce task-specific representations in the forward pass while reverse projections force the per-task gradients getting closer. However, learning parameters of the projections is a new optimization challenge. 
CosReg\cite{suteu2019cosreg} observes that well-performance models naturally obtains nearly orthogonal gradient angles. So they add a regular term into multi-task loss to modify gradients by back-propagation. We observe the similar phenomenon in STA without the regular term and multiple back-propagation as CosReg.

Multi-task learning obtains great success in different applications and problems. The most commonly used problem is Scene Understanding. It consists of both regression and classification tasks whose scales and training difficulties vary. We mainly use semantic segmentation and depth regression in this problem.
NYUv2~\cite{silberman2012nyuv2} and Cityscapes~\cite{cordts2016cityscapes} are most common datasets in multi-task learning thanks to their indoor and outdoor nature. Following related work, we evaluate our methods on both with different architectures.

Instance segmentation is a very important research area in the computer vision community. From Mask R-CNN\cite{he2017mask}, detection and semantic segmentation are trained together to improve the performance of instance segmentation. However, we find few multi-task works adopting on this application. We apply both STA and ISTA with Mask R-CNN, in order to show their applicability on instance segmentation.

\section{Method}

\subsection{Full Task (FT)}
% 介绍一下基本的Hard Parameter Sharing
In this section, we first introduce original multi-task learning. We follow hard parameter sharing \cite{caruana1997multitask}, an encoder-decoder architecture. For a $T$-task~($T\geq2$) learning problem, given a dataset $\mathcal{S} = \{(x_i, y_i^1, ..., y_i^T)\}_{i=1}^n$ with $x_i\in\mathbb{R}^d$ as a training example and $(y_i^1,...,y_i^T)$ as its $T$ annotations. Multi-task learning tries to train an encoder $f_{\mathcal{E}}$ with parameters $w_{\mathcal{E}}$ and $T$ task-specific decoders $\{f_\mathcal{D}^t\}_{t=1}^T$ with model parameters $\{w_\mathcal{D}^t\}_{t=1}^T$. During training, the encoder takes a batch of examples $\{x_i\}_{i=1}^m$ as inputs and provides sharing representations.  
The sharing representations are fed into each decoders to get $T$ predictions $\{f_{\mathcal{D}}^{t}(f_{\mathcal{E}}(x_i))\}_{t=1}^T$, respectively. 

Current deep learning methods use mini-batch gradient descent to optimize the model parameter. For hard parameter sharing, a mini-batch consists of several training examples and all of their annotations. For instance, a $m$ samples mini-batch contains $mT$ annotations~(Figure \ref{fig:overview}.a), the loss function on a mini-batch is defined as:
\begin{equation}
\mathcal{L}_{\mathrm{MTL}} = \frac{1}{m} \sum_{i=1}^m \sum_{t=1}^T \lambda_t \mathcal{L}_t(x_i, y_i^t).
\label{eq:loss_ma_batch}
\end{equation}
It is actually a convex combination of per-task losses with task-specific weights $ \{ \lambda_t \}_{t=1}^T $. Note that in this multi-task loss, we use full tasks to train each example. We call this way as Full Task~(FT), in order to distinguish from our methods: only use partial tasks for an example. During back-propagation, multi-task loss produces gradients which compete resources mainly on the encoder parameters, leading to the well-known optimization problems.

This work focuses on solving gradient conflicts. Prior works try to directly modify gradients in order to make them similar to each other, so that avoid cancelling when combining. However, the modified gradients lost the guarantee to point the optima of corresponding tasks, so that the multi-task model is easier to converge but may not improve for individual tasks. We are curious about whether there exists a way to reduce conflicts of gradients while not directly change gradients without meanings?

\subsection{Stochastic Task Allocation (STA)}
\label{sec:sa}
Previous works\cite{gradsurgery, suteu2019cosreg, chen2020gradrop} suggest that Full Task suffers from heavy gradient conflicts. To resolve this problem, they directly modify the gradients to reduce the elimination during joint learning. Their works successfully soften conflicts but do not account for an important problem: whether the new gradient update is an effective and accurate improvement for the optimization of individual tasks? There is obviously no guarantee. So in this work, we find a way to resolve gradient conflicts without explicitly changing gradients.

Looking closer to Full Task formulation, we suppose that training one example with full tasks supervised at a step may be the main factor of gradient conflicts. Different gradients of the same example compete for optimization resources in the encoder. 

With this insight, we introduce Stochastic Task Allocation~(STA), a mechanism that randomly allocates partial tasks for each example every step. STA proceeds as follows: (1)~In each step, for every example $x_i$ in a mini-batch $\mathcal{B}$, STA allocates a subset of full tasks $ST$ to it. (2)~For each $t$ in $ST$, it computes corresponding loss with loss function $\mathcal{L}_t$ and add up to task loss $l_t$. (3) With weights and mean per-task losses $ \{ l_t \}_{t=1}^T $, it allows us to compute the multi-task Loss $L_{MTL}$ and update the parameters of both encoder and decoders. The full procedure is described in \cref{alg:STA}.

This procedure, which is simple to implement, makes sure that each task loss is computed by minimal overlap example subsets, leading to the minimal competition on the example representation. Our experimental results show that it successfully reduces the gradient conflicts by implicitly concentrating gradient angles towards 90 degree. The results are shown in \cref{fig:cos_sa}. This experiment indicates that STA supports the hypothesis in CosReg and other related works\cite{suteu2019cosreg, zhao2018unrelated_module}. 
In \cref{sec:main_results}, we report our experimental results in details. The results demonstrate that our STA improves the performance in multi-task setting and outperform recent state-of-the-art methods.

\begin{algorithm}[tb]
\caption{STA Training Procedure} 
\label{alg:STA}
\DontPrintSemicolon
% Pre-define
    \KwInput{
        Number of Tasks $T$, 
        Encoder $f_{\mathcal{E}}(\cdot;w_\mathcal{E})$, 
        Decoders $\{ f_{\mathcal{D}}^t(\cdot;w_\mathcal{D}^t) \}_{t=1}^T$,
        Batch size $ m $,
        Dataset $\mathcal{S} = \{(x_i, y_i^1, \ldots, y_i^T)\}_{i=1}^N$,
        Task weights $ \{ \lambda_t \}_{t=1}^T $,
        Loss function $ \mathcal{L}_t $
    }
    \KwOutput{
        Encoder weights $ w_\mathcal{E} $, 
        Decoder weights $ \{ w_\mathcal{D}^t) \}_{t=1}^T $
    }

% Algorithm start.
    Randomly initialize $w_{\mathcal{E}}$ and $\{ w_{\mathcal{D}}^t \}_{t=1}^T$
    
    \For{$\mathrm{iteration}=0,1,\ldots$}
    {
        Sample a batch from Dataset $\mathcal{S}$: 
        $\mathcal{B} = \{(x_i, y_i^1,\ldots,y_i^T)\}_{i}^{m}$
        
        \For{$i=1,\ldots,m$}
        {
            Random Pick a subset of Task $ST \subset \{1,\ldots,T\}$ 
            
            \For{each $t \in ST~\mathrm{(in\,parallel)}$}
            {
                $l_t \gets l_t + \mathcal{L}_t (f_{\mathcal{D}}^{t}(f_{\mathcal{E}}(x_i)), y_i^t)$
            }
        }

        $ L_{MTL} \gets \frac{1}{T} \sum_{t=1}^T \lambda_t l_t $
        
        $ w_\mathcal{E} \gets w_\mathcal{E} - \eta \nabla_{w_\mathcal{E}} L_{MTL} $
        
        $ w_\mathcal{D}^t \gets w_\mathcal{D}^t - \eta \nabla_{w_\mathcal{D}^t} L_{MTL} $ For each $t$
    }
    
    \Return $ w_\mathcal{E}, \{ w_\mathcal{D}^t \}_{t=1}^T $ 
      
\end{algorithm}

\subsection{Interleaved Stochastic Task Allocation}
For STA, it only allocates part of tasks for an example. The rest of tasks may be picked in the next epoch. It is obviously inefficient for multi-task learning. In order to resolve this weakness, we introduce Interleaved Stochastic Task Allocation(ISTA) for a complimentary of STA in order to improve its data efficiency. Simply, ISTA uses a series of sub-iterations to allocate the rest of tasks. For each sub-iteration, ISTA samples part of the rest tasks for each example and computes losses like STA.

We provide detailed progress as follow: (1) ISTA initialize a set $RS_i$ as $ \{1,\ldots,T\} $ for each example $x_i$. It means all tasks can be allocated. (2) ISTA follows STA progress to allocate a part of tasks for each example. Instead, ISTA sample the subset from $RS_i$ rather than ${1,\ldots,T}$ in STA. (3) It removes $ST$ from $RS_i$ and computes corresponding losses. (4) At the end of a step, it adds up all task loss into multi-task loss and updates model parameters. (5) Repeat (2) to (4) until all $RS_i$ empty. Details are described in \cref{alg:ISTA}. Note that for most of our experiments, we set the size of $ST$ to 1. 

This procedure keeps STA-like training in each step while consecutively providing more adequate information in multi-task settings. In our experimental results, we show that ISTA always outperforms SA. Details locates in \cref{sec:ablation}.

\begin{algorithm}[tb]
\caption{ISTA Training Procedure} 
\label{alg:ISTA}
\DontPrintSemicolon
% Pre-define
    \KwInput{
        Number of Tasks $T$, 
        Encoder $f_{\mathcal{E}}(\cdot;w_\mathcal{E})$, 
        Decoders $\{ f_{\mathcal{D}}^t(\cdot;w_\mathcal{D}^t) \}_{t=1}^T$,
        Batch size $ m $,
        Dataset $\mathcal{S} = \{(x_i, y_i^1, \ldots, y_i^T)\}_{i=1}^N$,
        Task weights $ \{ \lambda_t \}_{t=1}^T $,
    }
    \KwOutput{
        Encoder weights $ w_\mathcal{E} $, 
        Decoder weights $ \{ w_\mathcal{D}^t) \}_{t=1}^T $
    }

% Algorithm start.
    Randomly initialize $w_{\mathcal{E}}$ and $\{ w_{\mathcal{D}}^t \}_{t=1}^T$
    
    \For{$\mathrm{iteration}=0,1,\ldots$}
    {
        Sample a batch from Dataset $\mathcal{S}$: 
        $\mathcal{B} = \{(x_i, y_i^1,\ldots,y_i^T)\}_{i}^{m}$
        
        $RS_i \gets \{ 1,\ldots,T \}$ For each $i \in \{1,\dots,m\}$  
        
        \Repeat 
        { $RS_i = \emptyset$ ~for each $i$ in $\{1,\ldots,m \}$}
        {
            
            \For{$i=1,\ldots,m$}
            {
                Random Pick a subset of Task $ST \subset RS_i$
                
                $RS_i \gets RS_i \setminus ST$
                
                \For{each $t \in ST~\mathrm{(in\,parallel)}$}
                {
                    $l_t \gets l_t + \mathcal{L}_t (f_{\mathcal{D}}^{t}(f_{\mathcal{E}}(x_i)), y_i^t)$
                }
            }

            $ L_{MTL} \gets \frac{1}{T} \sum_{t=1}^T \lambda_t l_t $
            
            $ w_\mathcal{E} \gets w_\mathcal{E} - \eta \nabla_{w_\mathcal{E}} L_{MTL} $
            
            $ w_\mathcal{D}^t \gets w_\mathcal{D}^t - \eta \nabla_{w_\mathcal{D}^t} L_{MTL} $ For each $t$
            }
        
    }
    
    \Return $ w_\mathcal{E}, \{ w_\mathcal{D}^t \}_{t=1}^T $ 
      
\end{algorithm}

\section{Experiments}
In this section, we first introduce experiment settings briefly. Then, we compare our methods with state-of-the-art methods from both gradient replacement and task weighting methods on three datasets, and evaluate the generalization of our methods on the instance segmentation with Mask R-CNN. Finally, we design several ablation studies to further analyze the proposed methods.

\subsection{Experiment Settings}

\noindent \textbf{Labeled Data.} Three MTL datasets are involved in experiments, including NYUv2\cite{silberman2012nyuv2}, Cityscapes\cite{cordts2016cityscapes} and COCO\cite{lin2014coco}. 

\noindent \textbf{Architectures.} For NYUv2, we examine on two architectures in MTL: SegNet~\cite{badrinarayanan2017segnet} following Uncertainty weights~\cite{kendall2018multi} and Resnet50\cite{he2016resnet} with deeplab-like heads~\cite{chen2018aspp} following MTI-Net\cite{vandenhende2020mti}. The former is commonly used in previous works and the latter is a modern network for segmentation. Details of architectures and training strategies are reported in supplement materials.

\begin{table}[t]
\begin{center}\resizebox{0.999\linewidth}{!}{
\begin{tabular}{cccc}
\toprule
  \multirow{2}{*}{Method}& Depth        & Segmentation & $ \Delta MTL $  \\
  & RMSE[m]($\downarrow$) & mIoU[\%]($\uparrow$) & [\%]($\uparrow$) \\
\midrule

\multirow{2}{*}{Single task}    & 0.747 & -      & + 0.00 \\
                                & -      & 54.71 & + 0.00 \\

\midrule
Full Task(FT)                           & 0.745 & 53.22 & - 0.74   \\
Uncertainty~\cite{kendall2018multi}     & 0.752 & 54.12 & - 0.30   \\
GradNorm~\cite{chen2018gradnorm}        & 0.753 & 54.09 & - 0.31   \\ 
DWA~\cite{liu2019mtan}                  & 0.745 & 53.80 & - 0.45  \\
MGDA~\cite{sener2018mtmo}               & 0.751 & 54.04 & - 0.33  \\
CosReg~\cite{suteu2019cosreg}           & 0.749 & 54.00 & - 0.35  \\
PCGrad~\cite{gradsurgery}               & 0.744 & 54.66 & - 0.02   \\
PCGrad+Uncertainty                      & 0.749 & 55.20 & + 0.24   \\
\midrule

STA             & 0.735         & 54.80    &  + 0.05     \\
STA+Uncertainty & 0.741         & 55.12    &  + 0.21     \\
ISTA             & 0.737 & 55.03       & + 0.16           \\
ISTA+Uncertainty & \textbf{0.734} & \textbf{56.03}  & \textbf{+ 0.66}\\
\bottomrule
\end{tabular}
}
\end{center}
\caption{Performance of recent MTL methods on NYUv2-13 using SegNet. Single task baselines show the performance of single tasks model. The Full Task means multi-task learning baseline with uniform weights. We bold the best performance and draw a box for the best performance of each task.}\label{table:nyuv2}
\end{table}

\subsection{Main Results}
\label{sec:main_results}
We first evaluate our methods on NYUv2 using different networks, including SegNet and Resnet50. The standard NYUv2 dataset\cite{silberman2012nyuv2} contains both depth and semantic segmentation labels (includes 894 classes) for a variety of indoor scenes \ie living room, bathroom, and kitchens \etc. NYUv2 is relatively small (795 training, 654 test images), but contains both regression and classification labels that have similar loss scales. Note that on semantic segmentation, we cluster 894 classes into 13 classes following \cite{liu2019mtan} for SegNet and 40 classes for Resnet50 following \cite{vandenhende2020mti}. For comparison, we use a different name for these two settings: NYUv2-13 and NYUv2-40.

SegNet results are reported in \cref{table:nyuv2}. We provide three metrics: mIoU for segmentation, RMSE for depth regression and $ \Delta MTL $ for comprehensive MTL comparisons, as ~\cite{vandenhende2020mti}.
STA outperforms most of the previous tasks and achieves 54.80\% mIoU which is comparable with single-task counterparts and highly superior to the Full task, suggesting that our task allocation leads to great performance indeed.
Meanwhile, ISTA reaches comparable performance on depth and higher mIoU on Segmentation than STA. It provides evidence that interleaved allocation may help performance than stochastic allocation.
Because task weighting methods do not inference gradient directions, they can be combined with our methods. As in \cref{table:nyuv2}, our methods with Uncertainty Weights reach 0.66 multi-task gains and the highest performance on both tasks in all methods like PCGrad and its combination with UW.

\begin{table}[ht]
\begin{center}\resizebox{0.999\linewidth}{!}{
\begin{tabular}{cccc}
\toprule
  \multirow{2}{*}{Method}& Depth        & Segmentation & $ \Delta MTL $  \\
  & RMSE[m]($\downarrow$) & mIoU[\%]($\uparrow$) & [\%]($\uparrow$) \\
\midrule

\multirow{2}{*}{Single task}    & 0.585 & -      & + 0.00 \\
                                & -      & 43.9 & + 0.00 \\

\midrule
Full Task(FT)                           & 0.587 & 44.4 & + 0.25   \\
Uncertainty~\cite{kendall2018multi}\dag     & 0.590 & 44.0 & + 0.05   \\
GradNorm~\cite{chen2018gradnorm}\dag        & 0.581 & 44.2 & + 0.15   \\ 
DWA~\cite{liu2019mtan}\dag                  & 0.591 & 44.1 & + 0.09  \\
MGDA~\cite{sener2018mtmo} \dag              & 0.576 & 43.2 & - 0.35  \\

\midrule
STA             & 0.583     & 45.0    &  + 0.55     \\
STA+Uncertainty & 0.579     & 44.8    &  + 0.45     \\

ISTA             &  0.584 &  45.7       &  \textbf{+ 0.90 }   \\
ISTA+Uncertainty &  0.578 &  45.3       &  + 0.70    \\
\bottomrule
\end{tabular}
}
\end{center}
\caption{Performance of recent MTL methods on NYUv2-40 using Resnet50 and deeplab-like ASPP module as decoders. We bold the best performance. Note that some results are imported from \cite{vandenhende2021survey}, we use \dag ~to sign them.}\label{table:nyuv2_resnet}
\end{table}

In Resnet50 experiments, we reproduce the results of \cite{vandenhende2020mti} and apply our methods on it following the same setting. All results report in \cref{table:nyuv2_resnet}. We observe different trends that most multi-task methods outperform single tasks except for MGDA. In this experiment, Our STA and ISTA still outperform all prior works and single-task models by a larger gap than SegNet results. 

We aim to use STA and ISTA to evaluate the outdoor scenarios, which may have heavier imbalance problem and different situations of gradient conflicts.
Cityscapes carries video frames shot in the streets of 50 urban cities with various annotations including disparity maps(use for depth regression), panoptic segmentation which can be split into 19 classes semantic segmentation and instance segmentation. The dataset contains nearly 5000 fine-annotated images with pixel-level labels and splits into 2975, 500, and 1500 for train, validation, and test respectively. 

As shown in \cref{table:city1}, STA outperforms Uncertainty Weights~(UW) and PCGrad in depth but drops a little on segmentation than UW. It suggests that this setting on Cityscapes suffers from huge imbalance problems. 
Meanwhile, ISTA reaches the best performance than related works and STA in both tasks: 3.720 for Depth and 66.26\% for segmentation.
With the same trend, STA and ISTA boost so much by uncertainty weights. ISTA with UW also achieves the best performance in this experiment with a huge gap than other methods.

\begin{table}[ht]
\begin{center}\resizebox{0.999\linewidth}{!}{
\begin{tabular}{cccc}
\toprule
  \multirow{2}{*}{Method}& Disparity        & Segmentation   & $ \Delta MTL $ \\
  & L1 Distance[px]($\downarrow$) & mIoU[\%]($\uparrow$) & [\%]($\uparrow$)\\
\midrule

\multirow{2}{*}{Single task}    & -      & 63.84  & + 0.00  \\
                                & 3.903  & -      & + 0.00    \\

\midrule
Full Task                               & 3.831 & 64.79  & + 0.51  \\
Uncertainty~\cite{kendall2018multi}     & 3.861 & 66.15  & + 1.18  \\
GradNorm~\cite{chen2018gradnorm}        & 3.718 & 63.54  & - 0.06  \\ 
PCGrad~\cite{gradsurgery}               & 3.846 & 64.28  & + 0.25  \\
DWA~\cite{liu2019mtan}                  & 3.842 & 64.14  & + 0.18  \\
MGDA~\cite{sener2018mtmo}               & 5.252 & 65.09  & - 0.05  \\

\midrule
STA             & 3.752           & 66.04       & + 1.17   \\
STA+Uncertainty & 3.821                    & 68.33  & + 2.29        \\

ISTA             & \textbf{3.720} & 66.26      & + 1.30    \\
ISTA+Uncertainty & 3.790          & \textbf{68.84}  & \textbf{+ 2.56}\\
\bottomrule
\end{tabular}
}
\end{center}
\caption{Performance of MTL methods on Cityscapes with the same architecture in \cref{table:nyuv2_resnet}. }\label{table:city1}
\end{table}

\subsection{Application to Instance Segmentation}
\label{sec:instance}
In this section, we use two datasets: Cityscapes and COCO~\cite{lin2014coco} with detectron2~\cite{wu2019detectron2} implementation for Mask R-CNN~\cite{he2017mask}. All hyper-parameters follow the default configuration in detectron2. All results are reported in \cref{table:Mask}.

For Cityscapes, we use two tasks as supervised signals: detection and semantic segmentation. Origin Mask R-CNN evaluates the instance segmentation, so we only report instance segmentation and detection metrics. Note that we do not utilize Uncertainty Weights and GradNorm in Mask R-CNN experiments because it is hard to make UW and GN work on detection models and this topic is out of this work. 

However, we also found that STA and ISTA achieve superior performance compared to FT baseline by a clear margin. 
Mask R-CNN suggests that models trained on cityscapes may suffer from overfitting problems. We observe the same phenomenon, so we only train half steps for STA and ISTA. The results show that our methods still boost multi-task performance. In conclusion, ISTA surpasses SA by a 1 to 1.5\% margin on both tasks. 

COCO is a large dataset with about 118K training and 5K validation images. By evaluating on COCO, we aim to show that our methods can generalize to huge datasets. As shown in \cref{table:Mask}, STA and ISTA still surpass the baseline. However, STA performs better than ISTA which is totally different from previous experiments. We leave it for future works.

\begin{table}[ht]
\begin{center}\resizebox{0.999\linewidth}{!}{
\begin{tabular}{ccccccc}
\toprule
  \multirow{2}{*}{Dataset} & \multirow{2}{*}{Method}& \multicolumn{2}{c}{Instance Segmentation} & \multicolumn{2}{c}{Detection} & $ \Delta MTL $\\ 
  \cmidrule(r){3-4} \cmidrule(r){5-6}
  &  & AP & AP50  & AP & AP50 & [\%]($\uparrow$)\\
\midrule

  \multirow{3}{*}{Cityscapes}& Baseline    & 36.73 & 62.56  & 41.67 & 65.06 & + 0.00  \\

  & STA    & \textbf{37.93} & 64.99  & 42.50 & 67.33 & + 4.43 \\
  & ISTA  & 37.85 & \textbf{65.86}  & \textbf{42.78} & \textbf{68.02} & + 4.87 \\
\midrule
  \multirow{3}{*}{COCO}&Baseline    &  38.624 & 59.428   & 35.206 & 56.559 & + 0.00 \\

  & STA     & \textbf{39.334} & \textbf{60.312}  & \textbf{35.763} & \textbf{57.291} & +0.72 \\
  & ISTA  & 38.994 & 59.699  & 35.497 & 56.729 & + 0.28 \\

\bottomrule
\end{tabular}
}
\end{center}
\caption{Mask R-CNN on Cityscapes and COCO}\label{table:Mask}
\end{table}

\subsection{Ablation Study}
\label{sec:ablation}
In the ablation study, the goals of our experiments are: 1) we first compare with Full Task(FT), STA and ISTA, in order to show progressive improvements of this work. 2) we claim that STA and ISTA may require more steps than Full Task. It is clear that our methods still surpass FT which has the same amount of steps. 3) Because task weighting methods do not change the direction of gradients, our methods can transparently combine with task weighting methods like Uncertainty weights and GradNorm. We give more details about it. 4) PCGrad modifies the gradients in a different way than our methods. We are curious about the applicability of the combination. 5) ISTA progressively allocates tasks for a batch of same examples while STA meets the same examples after a long time. There is an interesting question: if we manually reduce the time gap, is it a benefit for performance?

\noindent \textbf{Full Task vs Partial Task.}
Firstly, we aim to show the superior of our methods which use only partial tasks for each example compared to the origin Full Task MTL. As shown in \cref{table:nyuv2}, \cref{table:city1} and \cref{table:Mask}, STA and ISTA surpass Full Task by a large margin whether they combines with Uncertainty Weights or not. 

\begin{figure}[!t]
    \centering
    \includegraphics[width=\linewidth]{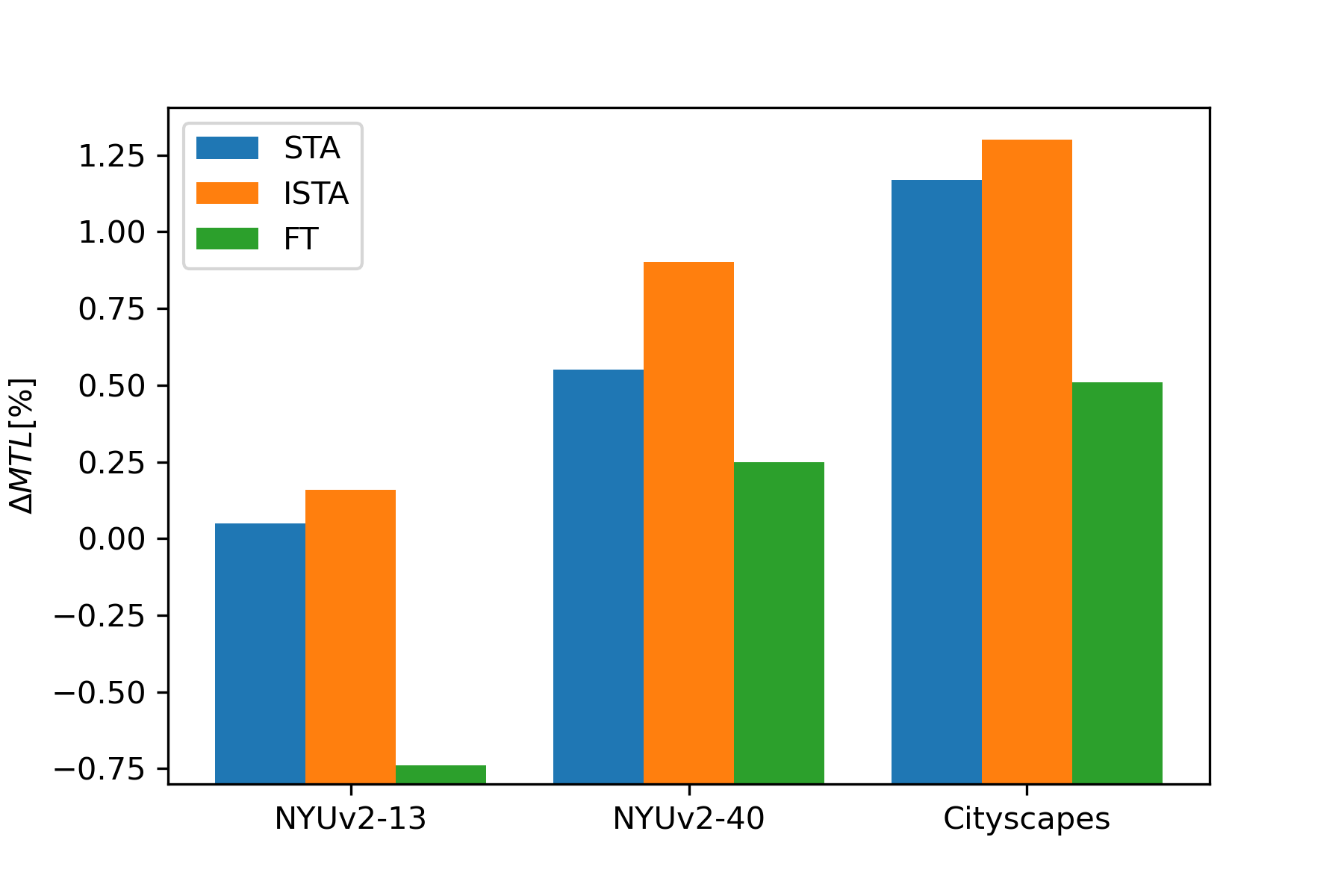}
    \caption{We plot $ \Delta MTL$ on all experiments to show that STA and ISTA outperforms origin Full Task. Further, ISTA always perform better than STA.}
    \label{fig:ablation1}
\end{figure}

For comparison, we summary results from all experiments in \cref{fig:ablation1}. In most situations, ISTA seems more general on multi-task performance empirically except for COCO datasets: it reaches the highest performance on both Cityscapes and NYUv2 not only on part of tasks but on both tasks and various network architectures. 

\noindent \textbf{More steps is not the key.}
For the results in NYUv2-13 and Cityscapes, we run all experiments for the same epoch. However, STA and ISTA only use part of tasks for one step so that the length of an epoch increases resulting in growing steps. In order to exclude the factor of far more steps, we run Full Task experiment with the same step as our methods, short as FT+. We report the results in \cref{table:equal_step}. 
FT+ gains about 0.3\% improvements on $ \Delta MTL $, while STA and ISTA reach about 0.7\% and 0.9\% respectively, suggesting that the task allocation improves multi-task performance mainly.

In Mask R-CNN on Cityscapes and NYUv2-40 experiments, STA and ISTA run for the same steps while also outperforming baseline by a margin, as in \cref{table:Mask}. It indicates that STA and ISTA may need different optimization schedules from Full Task. We assume that distinct batch size setting of heads and backbones leads to this schedule difference. The same phenomenon is observed by IMTL\cite{liu2020imtl} which employs different updating strategies to the parameters of encoder and decoders respectively. 

\begin{table}[ht]
\begin{center}\resizebox{0.999\linewidth}{!}{
\begin{tabular}{ccccc}
\toprule
  \multirow{2}{*}{Dataset}& \multirow{2}{*}{Method} & Depth        & Segmentation & $ \Delta MTL $  \\
  & & RMSE[m]/L1[px]($\downarrow$) & mIoU[\%]($\uparrow$) & [\%]($\uparrow$) \\

\midrule

\multirow{4}{*}{NYUv2-13} & Full Task(FT)   & 0.745 & 53.22 & - 0.74   \\
& FT+             & 0.748 & 53.69 & - 0.51   \\

\cmidrule(r){2-5}
& STA             & 0.735 & 54.80    &  + 0.05     \\
& ISTA            & 0.737 & 55.03       & + 0.16           \\

\midrule
\multirow{4}{*}{Cityscapes} & Full Task(FT)   & 3.831 & 64.79  & + 0.51   \\
& FT+             & 3.757 & 65.29 & + 0.80   \\

\cmidrule(r){2-5}
& STA             & 3.752 & 66.04       & + 1.17     \\
& ISTA            & 3.720 & 66.26       & + 1.30     \\
\bottomrule
\end{tabular}
}
\end{center}
\caption{We show the results of STA, ISTA, FT and FT+ which trains for the same time as our methods. We claim that although FT+ improves, it still falls behind STA and ISTA.}\label{table:equal_step}
\end{table}

\begin{figure}[!t]
    \centering
    \includegraphics[width=\linewidth]{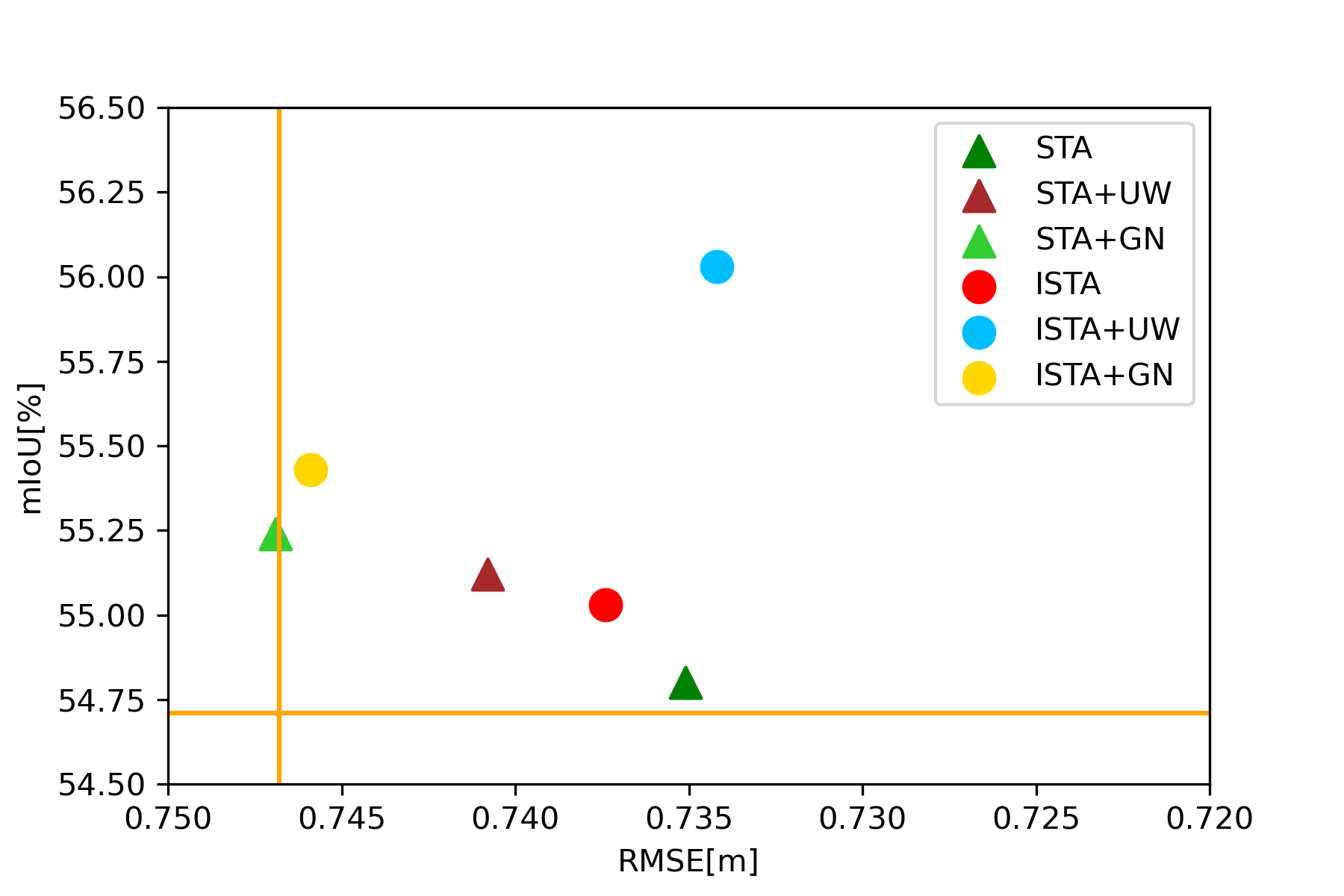}
    \caption{An overview of the combination between our methods, UW and GN on NYUv2-13}
    \label{fig:nyuv2_tw}
\end{figure}

\begin{figure}[!t]
    \centering
    \includegraphics[width=\linewidth]{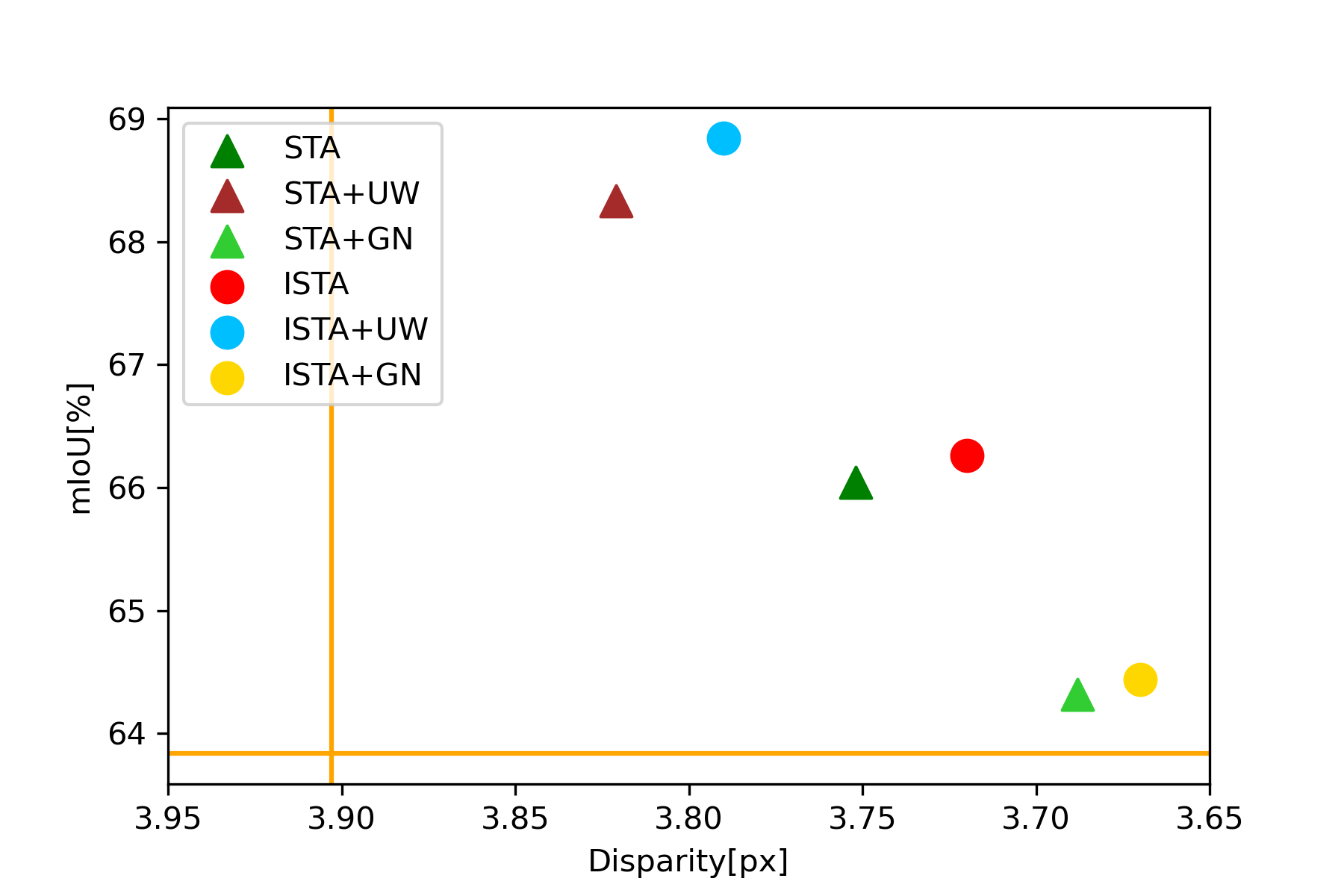}
    \caption{An overview of the combination between our methods, UW and GN on Cityscapes}
    \label{fig:city_tw}
\end{figure}

\noindent \textbf{Combination with Task Weighting.}
As we mentioned before, task weighting methods do not change the directions of gradients, so it is obvious that we can combine our methods with task weighting. We evaluate the experiments on both NYUv2-13 and Cityscapes. Results are plotted on \cref{fig:nyuv2_tw} and \cref{fig:city_tw}.

For NYUv2-13, ISTA with Uncertainty weights performs best by a clear gap. For cityscapes, UW stills provides large improvements with both STA and ISTA, while GN is not ideal for segmentation but reaches the highest depth accuracy.
In conclusion, the results suggest that our methods with task weighting can further improve multi-task performance.

\noindent \textbf{Gradient Projection with STA.}
PCGrad projects per-task gradients to make them closer in the angle while STA and CosReg prefer more uncorrelated gradients. As shown in \cref{fig:cos_sa}, we observe that our STA exactly push the distribution of included angles to 90 degree. However, there still exists a small part of angles larger than 90. By the way, we are curious about the problem: whether STA combined with PCGrad can help multi-task learning further?

We provides experiments on Cityscapes in \cref{table:pcgrad+sa}. The results indicate that the combination improves performance compared to PCGrad. However, it plays a negative role in both STA and ISTA: the multi-task performance drops by about 0.3-0.4\%. 

In conclusion, we claim: 1) uncorrelated gradients always improve the model performance. 2) it is not essential or even hurt to project gradients while they are already close to 90 degree. The results support our concern about the works of gradient modification.

\begin{table}[ht]
\begin{center}\resizebox{0.999\linewidth}{!}{
\begin{tabular}{cccc}
\toprule
  \multirow{2}{*}{Method}& Disparity        & Segmentation   & $ \Delta MTL $ \\
  & L1 Distance[px]($\downarrow$) & mIoU[\%]($\uparrow$) & [\%]($\uparrow$)\\
\midrule

PCGrad~\cite{gradsurgery}   & 3.846    & 64.28       & + 0.25  \\
STA                         & 3.752    & 66.04       & + 1.17   \\
ISTA                        & 3.720    & 66.26       & + 1.30   \\
STA+PCGrad                  & 3.781    & 65.43       & + 0.86   \\
ISTA+PCGrad                 & 3.772    & 65.52       & + 0.90   \\
\bottomrule
\end{tabular}
}
\end{center}
\caption{Performance of MTL methods on Cityscapes with the same architecture in \cref{table:nyuv2_resnet}. }\label{table:pcgrad+sa}
\end{table}

\noindent \textbf{Interpolation between STA and ISTA.}
ISTA is a special variation of STA, where it manually allocates the rest of tasks progressively over a consecutive time. STA also meets the same example with other tasks after a random time. We manually control the time gap between STA meets two tasks of the same example. In detail, at step $t$, a batch of examples is trained by the model with task allocations $\{{ST}_i\}_{i=1}^m$. At step $t+gap$, the same batch of examples is learned again with another allocation $\{\widehat{ST}_i\}_{i=1}^m$ which are randomly chosen. ISTA is a special case where the gap is 1. The goal of this experiment is to find if there exits a performance boost when the time gap changes. We design the experiments on NYUv2-13 and report the results in \cref{table:interp}.

The results show that the interpolation does not inference much. STA with different gaps performs similarly except for ISTA: the gap equals 1. In a word, we suppose that the key point of ISTA may be the consecutive allocation. 

\begin{table}[ht]
\begin{center}\resizebox{0.999\linewidth}{!}{
\begin{tabular}{cccc}
\toprule
  \multirow{2}{*}{Method}& Depth        & Segmentation & $ \Delta MTL $  \\
  & RMSE[m]($\downarrow$) & mIoU[\%]($\uparrow$) & [\%]($\uparrow$) \\
\midrule

ISTA(Gap=1)      & 0.737        & 55.03    & + 0.16     \\
STA-Gap=2        & 0.738        & 54.84    & + 0.07     \\
STA-Gap=4        & 0.736        & 54.76    & + 0.03     \\
STA-Gap=8        & 0.736        & 54.66    & - 0.01     \\
STA-Gap=16       & 0.737        & 54.77    & + 0.04     \\
STA              & 0.735        & 54.80    & + 0.05     \\

\bottomrule
\end{tabular}
}
\end{center}
\caption{Experiment results of interpolating the time gaps between STA and ISTA. }\label{table:interp}
\end{table}

\section{Limitations and Ethical Discussion}
In this work, we discover a novel, sampling solution to gradient conflict problems by allocating random partial tasks for each example. In this process, we do not study how to choose tasks but randomly sample from a task set. Task grouping methods~\cite{strezoski2019task_group1, alonso2016task_group2} are one of the topics in multi-task learning but less related to gradient conflicts, so we leave it for future works. 
Additionally, it seems not necessary to train a multi-task model with full-annotated examples. For example, NYUv2 has 795 full-annotated images while there are 407K images with only depth labels. If these 407K images can be used for training, it is possible to improve multi-task performance further. This work does not include these experiments but can be a good step.

For ethics, we carefully review the open-source datasets used in this work. In NYUv2, all annotations are labeled automatically and no personally identifiable information has been used in labels. In Cityscapes and COCO, we abide by the term of use. 
In addition, this work has no foreseeable negative societal impact.

\section{Conclusion}
In this work, we first introduce Stochastic Task Allocation(STA), a novel method to resolve gradient conflicts by implicitly concentrating gradient angles towards 90 degree without any gradient modification. Our empirical results indicate that STA surpasses state-of-the-art task weighting and gradient modification methods. We demonstrate that STA is general to multiple datasets and architectures in the multi-task context. Further, we propose ISTA as a complementary of STA to improve its data efficiency by allocating the rest of tasks until all tasks have been used. We empirically show that ISTA outperforms STA and achieves the best multi-task performance on most datasets especially when combined with uncertainty weights. 

While we study multi-task supervised learning in this work, we suspect the gradient conflicts also exist in other settings, such as reinforcement learning, meta-learning, and multi-object optimization. Due to the simplicity and its intuitive insights, we expect that STA can be expanded into this setting. Further, a multi-task dataset is hard to build thanks to the high cost of annotation. STA points a way that there may be no need to annotate full tasks for all examples, so that more datasets with partial annotations may be utilized by the multi-task community. We hope this work can be a start point of these challenges.

%%%%%%%%% REFERENCES
{\small

\bibliographystyle{ieee_fullname}
}

\end{document}